# Parking spot classification based on surround view camera system

Andy Xiao*[a], Deep Doshi*[a], Lihao Wang*[a], Harsha Gorantla[a], Thomas Heitzmann[a], and Peter Groth[a]

[a]Dept. of R&D, Comfort Driving and Assistance, 2121 S El Camino Real, San Mateo, CA US 94403

## ABSTRACT

Surround-view fisheye cameras are commonly used for near-field sensing in automated driving scenarios, including urban driving and auto valet parking. Four fisheye cameras, one on each side, are sufficient to cover 360° around the vehicle capturing the entire near-field region. Based on surround view cameras, there has been much research on parking slot detection with main focus on the occupancy status in recent years, but little work on whether the free slot is compatible with the mission of the ego vehicle or not. For instance, some spots are handicap or electric vehicles accessible only. In this paper, we tackle parking spot classification based on the surround view camera system. We adapt the object detection neural network YOLOv4 with a novel polygon bounding box model that is well-suited for various shaped parking spaces, such as slanted parking slots. To the best of our knowledge, we present the first detailed study on parking spot detection and classification on fisheye cameras for auto valet parking scenarios. The results prove that our proposed classification approach is effective to distinguish between regular, electric vehicle, and handicap parking spots.

## 1. INTRODUCTION

Within urban areas, parking has become a major source of frustration for drivers. The search for parking spaces has become more arduous than ever [1], with motorists spending an average of 17 hours each year searching for parking on streets, lots, and garages. According to INRIX's analysis [2], this hunt results in approximately $345 of wasted time, fuel, and emissions per driver. To address this issue, developing parking solutions that save time and money for drivers and fleets, while also reducing congestion and pollution in urban areas, has become essential. Among the various efforts aimed at tackling this issue, parking spot detection is a pivotal component. By implementing efficient parking spot detection technologies, drivers can be guided towards available spaces quickly, reducing the time and resources wasted on fruitless searches.

Two primary approaches are commonly used for the detection of empty parking spots. The first method involves installing sensors in each parking spot to detect the presence of a car. While this approach is effective, it can be quite costly to implement [3]. Alternatively, the second approach relies on visual information, often obtained from pre-installed security cameras, to determine whether a parking lot is empty or occupied. Both approaches depend on the factor of infrastructure availability. Leveraging existing onboard sensors installed on vehicles presents a promising opportunity to eliminate the dependency on additional infrastructure for parking spot detection.

There are two main methods to employ onboard sensors for detection of parking spots: the free-space-based parking spot detection method and the vision-based parking spot detection method [4]. The free-space-based method primarily relies on the reflection principle of ranging sensors, such as ultrasonic, radar, and laser scanner. This method assesses the space between adjacent vehicles to determine the size and location of the target parking space. However, the free-space-based method has certain limitations. It necessitates the presence of adjacent vehicles around the target parking space for effective detection, and its accuracy is contingent upon the poses of these adjacent vehicles. Additionally, this method suffers from drawbacks such as a limited detection range, potential blind spots in detecting parking spaces. To address the limitations of the free-space-based method, researchers are increasingly focusing on the vision-based method [5-10]. In the vision-based method, parking spots are determined by identifying and locating the parking line segments drawn on the ground. This method's performance is not reliant on the presence or posture of adjacent vehicles, making it more flexible in various parking scenarios. In addition, detecting markings in the vision-based method enables the identification of various parking spot categories, including electric vehicle (EV) charging spots and handicap-accessible parking spots.

*These authors contributed equally. Corresponding author contact info: andy.xiao@valeo.com; phone +1 650 541-6955;

Vision-based parking space detection methods can be broadly categorized into two distinct groups: traditional methods and deep learning-based methods [10]. Traditional methods rely on classical computer vision techniques and algorithms to detect and process visual features in the parking environment. Examples of traditional methods include Hough Transform for line segment detection, corner detection algorithms, and other image processing techniques. These methods often involve manual feature engineering and geometric reasoning to identify parking spots [4]. Unlike traditional methods, deep learning-based methods leverage neural networks and adopt an end-to-end learning approach, without the need for hand-crafted feature engineering as reported in [6-10, 14]. These methods achieve high accuracy in detecting parking spaces, even in complex and diverse parking lot scenarios. Deep learning-based methods typically detect the marking points or critical features of the parking space (e.g. intersection points of parking lines) by using object detection neural networks [4], like YOLO [15], SSD [16], YOLOv2 - v4 [13, 17, 18], etc. In reference [6], the authors proposed a parking space detection method based on YOLOv2. [14] introduced a vacant parking space detection method called VPS-Net, which utilizes YoloV3 object detection algorithm. In reference [12], we proposed a parking slot detection method, the Holistic Parking Slot Network (HPS-Net) based on custom adaptation of the YOLOv4 algorithm.

As mentioned, extensive research has been conducted on parking slot detection using onboard cameras, and with a primary focus on determining the occupancy status. However, a crucial aspect that has been largely overlooked is identifying whether a free parking slot is compatible with the needs of the ego vehicle, such as being handicap-accessible or designated for electric vehicles. This paper addresses the crucial matter of parking spot classification using the surround view camera system (SVS). Building upon our previous work [12] on parking slot detection, this paper presents an extension of HPS-Net for comprehensive slot classification. Our proposed approach is able to detect and classify parking slots into various categories, including regular, handicap-accessible, and electric vehicle (EV) charging spots. To the best of our knowledge, our work represents the pioneering effort in addressing parking slot classification.

The remaining sections of this paper are structured as follows: Section 2 provides an overview of our proposed system for parking spot detection and classification. Section 3 presents a comprehensive explanation of the methodology employed in this research. Section 4 presents the experimental results and includes a detailed discussion of the findings. Finally, in Section 5, we conclude the paper by summarizing the main contributions and discussing potential areas for future research.

## 2. SYSTEM OVERVIEW

The diagram presented in Fig. 1 outlines the pipeline of our proposed parking spot detection and classification system. It consists of two main modules: the surround-view synthesis part and the parking-slot detection and classification part. During operation, the system first employs the surround-view synthesis module to create a holistic view from the inputs of the four fisheye cameras. Subsequently, the adapted YOLOv4 neural network is used to identify and categorize parking spots.

Fig.1. Diagram illustrating the full pipeline of the proposed parking spot detection and classification system.

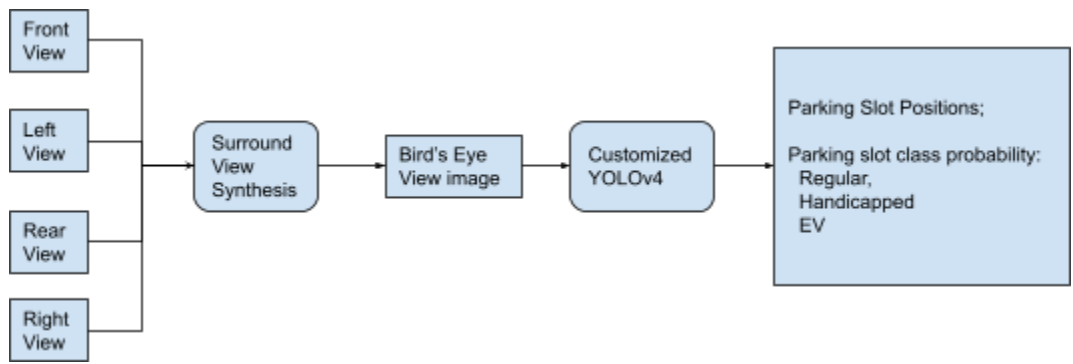

## 3. METHODOLOGY

### 3.1 Inverse Perspective Mapping

In this parking spot detection and classification system, four surround-view cameras are employed to achieve comprehensive coverage. The camera configuration, as depicted in Fig. 2, is a standard setup commonly found in high-configuration commercial cars. This arrangement consists of four fisheye cameras which can capture a wide-angle view of their respective surroundings, enabling 360° parking spot detection and classification.

The intrinsic and extrinsic parameters of each camera are calibrated offline. Each pixel in the camera image is then projected onto the ground plane with respect to the vehicle's coordinate system. This process, also known as Inverse Perspective Mapping (IPM), allows us to transform the camera view into a bird's-eye view perspective [7-9, 11, 19]. The process begins by rectifying the fisheye images to eliminate lens distortion [20]. Subsequently, the projection operation is conducted as follows:

$$[u, v, 1]^t \simeq H_{3\times 3} \cdot [x, y, 1]^t \tag{1}$$

where *[u, v]* represents the 2D coordinates on the bird's eye view plane in the scene, while *[x, y]* corresponds to the pixel coordinates in the image plane. The transformation between these two coordinate systems is described by the homography matrix $H_{3\times 3}$. The homography matrix is estimated using the intrinsic parameters of the camera and the extrinsic parameters [21]. After the inverse perspective projection, a large bird's eye view image can be built from four fisheye images, as shown in Fig. 3.

### 3.2 Parking Spot Detection and Classification

Inspired by [12], we model each parking slot using a four-point polygon (i.e. quadrilateral). The four vertices (or corners) are arranged in a specific sequence of entrance-left, entrance-right, ending-pleft, ending-right, as follows:

$$\{(x1, y1), (x2, y2), (x3, y3), (x4, y4)\} \tag{2}$$

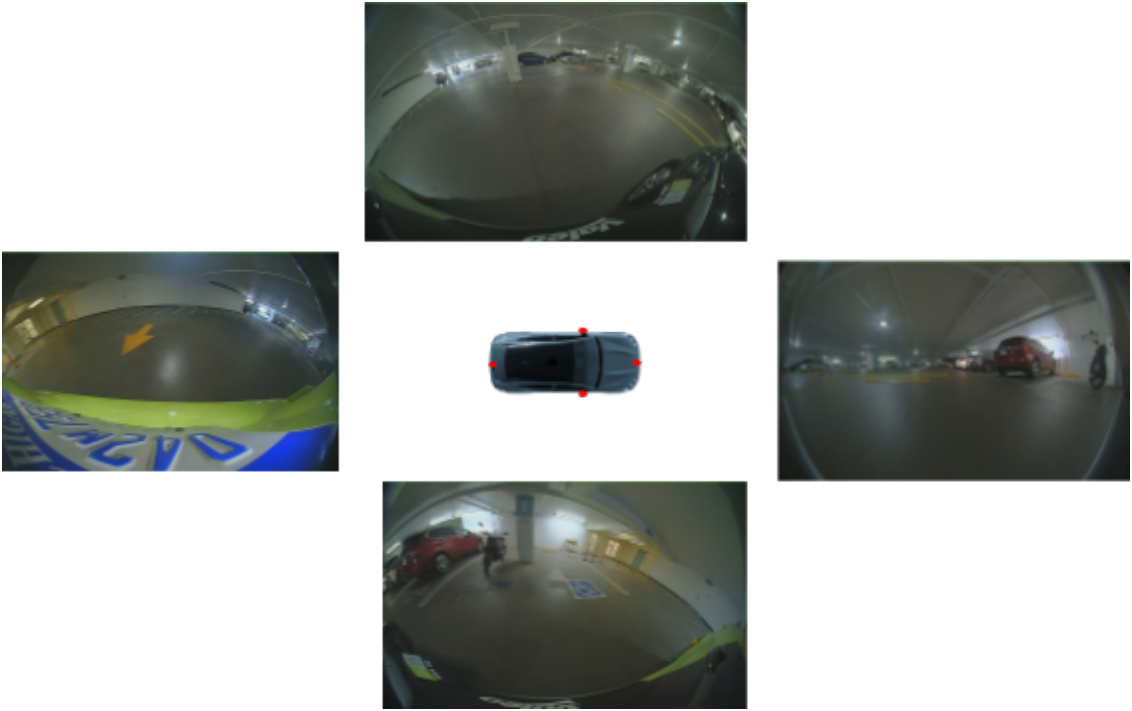

Fig. 2. Configuration of four surround-view cameras used in parking spot detection.

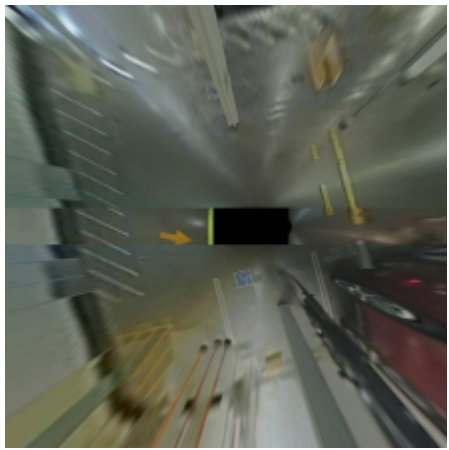

Fig. 3. The synthetic bird's eye view image from four surround-view cameras using IPM

One primary advantage of this representation is that it can fit to all possible shapes of parking spaces, including perpendicular, parallel, and fishbone, as shown in Fig. 4, making our approach highly general and applicable to various scenarios.

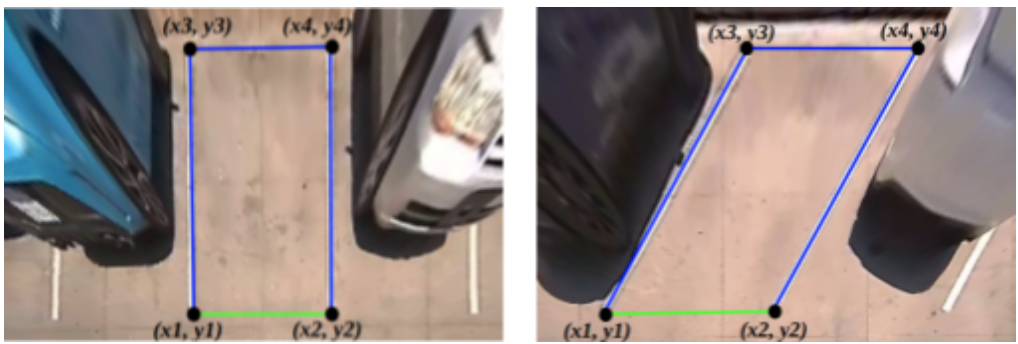

Fig. 4. Polygon parking model for perpendicular and fishbone cases. Entrance lines are marked in green.

For parking classification, we focus on three most common categories: regular, handicapped, and EV. Fig. 5 illustrates examples from each category.

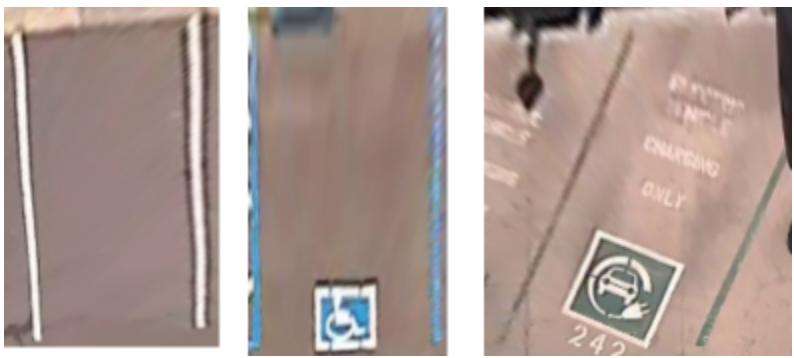

Fig.5. Different parking classes (from left to right: regular, handicapped, and EV).

Loss functions play a crucial role in deep learning models. To ensure that our model can converge during the training process, we employ a specific IoU (Intersection over Union) loss for polygon corners regression and binary cross-entropy loss for classification. Other loss components such as objectness follow YOLOv4 [13]. As a result, the total loss can be expressed as follows:

$$L_{total} = L_{polygon\_IoU} + L_{bce} + L_{obj} \qquad (3)$$

## 4. EXPERIMENT AND DISCUSSION

### 4.1 Dataset

To facilitate the training and evaluation of our proposed parking slot detection and classification approach, we have established a comprehensive benchmark dataset of surround view images. The images in this dataset were collected from various indoor and outdoor parking sites, representing diverse real-world scenarios. The dataset comprises three distinct categories of parking spots, including Regular, Handicap, and Electric-Vehicle (EV), as shown in Fig. 5.

The fisheye image resolution is 1920 × 1080, capturing a wide-angle view of the surroundings. Through the synthesis process, we generate a bird's eye view image with a resolution of 1024 × 1024. This synthesized image covers a 25m × 25m flat surface, providing a detailed representation of the environment from a top-down perspective.

The dataset (train and test) consists of 1352 generated bird's eye view images, encompassing a total of 11,505 labels distributed across three categories, as shown in Table 1.

Table 1. The label distribution for the parking slot detection and classification dataset.

| Class | Train set | Test set | Total |
|---|---|---|---|
| Regular | 8846 | 1808 | 10654 |
| Handicapped | 429 | 104 | 533 |
| EV | 280 | 38 | 318 |

### 4.2 Parking Slot Detection and Classification Performance

We have conducted a comprehensive evaluation of our method on the test set, assessing its performance in both challenging indoor and outdoor environments. To visually showcase the results of detection and classification, we present qualitative examples in Figures 6, 7, and 8. In each figure, the left image corresponds to the ground truth (GT) label, while the right image shows our model's prediction.

In Figure 6, we observe that all four handicapped spots are successfully detected and correctly classified, which demonstrates the robustness of our approach in identifying specialized parking spaces. However, we also notice that there is a missed detection of the regular spot in the top right corner, revealing a limitation in our current implementation.

The results in Figures 7 and 8 provide additional insights into our method's performance in outdoor environments. In Figure 7, we observe a missing detection for the EV parking spot in the top left. Similarly, in Figure 8, we identify a misclassification where a handicapped spot is incorrectly identified as a regular spot in the bottom left.

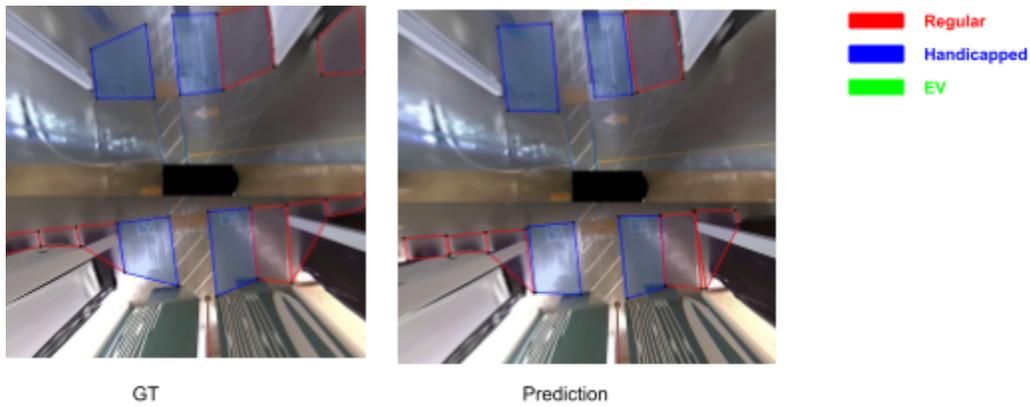

Fig.6. Examples of parking lot detection and classification results in a challenging indoor environment. Four handicapped spots are detected and classified correctly, despite a missed detection of the regular spot in the top right corner.

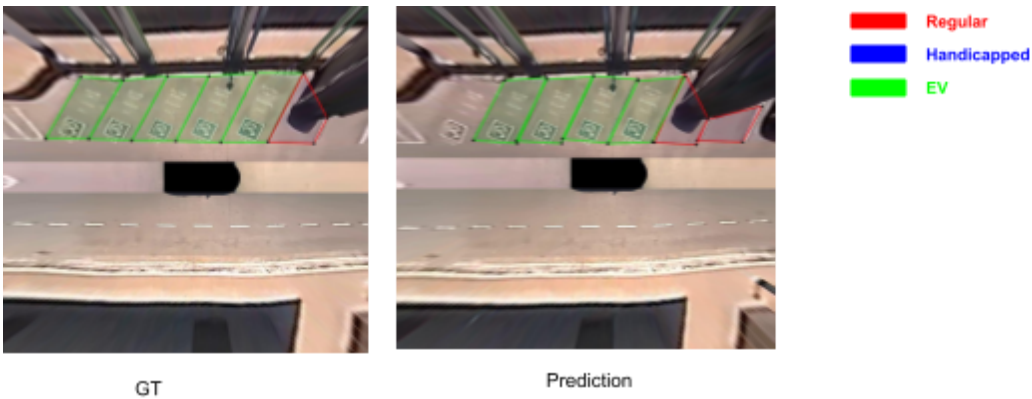

Fig.7. Examples of parking lot detection and classification results in an outdoor environment. Four out of the five EV spots are accurately identified, with a missing detection of an EV spot in the top left.

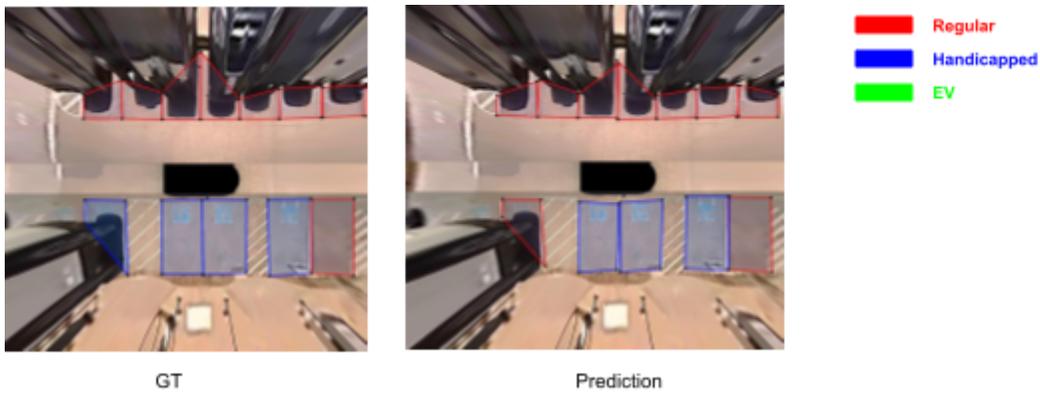

Fig.8. Examples of parking lot detection and classification results in an outdoor environment. All the regular spots are accurately detected and correctly classified. In the bottom left a handicapped spot is misclassified as a regular one.

To quantitatively assess the performance of our parking spot detection and classification results, we utilize essential metrics such as Precision, Recall, F-1 score, and Average Precision (AP) [12, 15]. The testing results for each metric are presented in Table 2. We observe that the regular parking spots achieve the highest AP at an Intersection over Union

(IoU) of 0.5, outperforming the other two classes. The mean AP over the three classes at an IoU threshold of 0.5 is 0.656. The results validate the performance of our parking spot detection and classification approach and demonstrate its ability to accurately identify and classify parking spots in various real-world scenarios.

Table 2. Parking slot detection and classification testing results.

| Class | Precision | Recall | F1 | AP @ 0.5 |
|---|---|---|---|---|
| Regular | 0.619 | 0.866 | 0.722 | 0.762 |
| Handicapped | 0.520 | 0.594 | 0.555 | 0.496 |
| EV | 0.742 | 0.710 | 0.726 | 0.710 |
| All | 0.627 | 0.724 | 0.668 | 0.656 |

## 5. Conclusion

In this paper, we present a novel approach for parking slot detection and classification by adapting YOLOv4, an advanced object detection neural network. Our method makes use of four surround view cameras integrated into the vehicle, capturing images that are later transformed into a bird's-eye view using inverse perspective mapping. The neural network is then employed to detect parking slots, classifying them into three categories: regular, handicapped, and EV. The effectiveness of our proposed method is rigorously validated through experiments, showcasing a mean average precision of 0.656 at an IoU threshold of 0.5.

In the future, our primary focus for further advancements is to leverage the four fisheye cameras as the input for the neural network. We aim to eliminate the need for IPM, streamlining the entire process and potentially enhancing the overall performance of the parking slot detection and classification system. Additionally, we also plan to integrate the neural network with multi-task capabilities, enabling dynamic object detection such as vehicle and pedestrian detection, thereby expanding its applicability.

## References

[1] Weinberger, R., Millard-Ball, A., Fabusuyi, T. et al. , Parking Cruising Analysis Methodology Project Report, U.S. Department of Transportation Federal Highway Administration, FHWA-HOP-23-004, March 2023.
[2] Graham Cookson, The Impact of Parking Pain in the US, UK and Germany, 2017, https://inrix.com/wp-content/uploads/2017/07/INRIX_Parking_Pain_Infog_US_HR.pdf
[3] Li W, Cao H, Liao J, Xia J, Cao L, Knoll A. Parking Slot Detection on Around-View Images Using DCNN. Front Neurorobot. 2020 Jul 24;14:46.
[4] Ma Y, Liu Y, Shao S, Zhao J and Tang J. (2022). Review of Research on Vision-Based Parking Space Detection Method. International Journal of Web Services Research. 19:1. (1-25).
[5] Xu, J., Chen, G., & Xie, M. (2000). Vision-guided automatic parking for smart car. Proceedings of the IEEE Intelligent Vehicles Symposium 2000, 725-730.
[6] L. Zhang, J. Huang, X. Li and L. Xiong, "Vision-Based Parking-Slot Detection: A DCNN-Based Approach and a Large-Scale Benchmark Dataset," in IEEE Transactions on Image Processing, vol. 27, no. 11, pp. 5350-5364, Nov. 2018.
[7] Suhr J and Jung H. (2022). End-to-End Trainable One-Stage Parking Slot Detection Integrating Global and Local Information. IEEE Transactions on Intelligent Transportation Systems. 23:5. (4570-4582).
[8] Chen Z, Qiu J, Sheng B, Li P and Wu E. (2021). GPSD: generative parking spot detection using multi-clue recovery model. The Visual Computer: International Journal of Computer Graphics. 37:9-11. (2657-2669).
[9] Yu Z, Gao Z, Chen H and Huang Y. SPFCN: Select and Prune the Fully Convolutional Networks for Real-time Parking Slot Detection. 2020 IEEE Intelligent Vehicles Symposium (IV). (445-450).

[10] Z. Wu, W. Sun, M. Wang, X. Wang, L. Ding and F. Wang, "PSDet: Efficient and Universal Parking Slot Detection," 2020 IEEE Intelligent Vehicles Symposium (IV), Las Vegas, NV, USA, 2020, pp. 290-297.

[11] Zizhang Wu, Yuanzhu Gan, Xianzhi Li, et al., "Surround-View Fisheye BEV-Perception for Valet Parking: Dataset, Baseline and Distortion-Insensitive Multi-Task Framework", IEEE Transactions on Intelligent Vehicles, vol.8, no.3, pp.2037-2048, 2023.

[12] Wang L , Musabini A, Leonet C, Benmokhtar R, Breheret A, Yedes C, Bürger F, Boulay B, Perrotton X. (2023). Holistic Parking Slot Detection with Polygon-Shaped Representations. 2023 IEEE/RSJ International Conference on Intelligent Robots and Systems (IROS), Detroit, MI, USA, 2023.

[13] Bochkovskiy A., Wang C.-Y., Liao H.-Y. M., "Yolov4: Optimal speed and accuracy of object detection," 2020. [Online]. Available: https://arxiv.org/abs/2004.10934

[14] Li W, Cao L, Yan L, Li C, Feng X, Zhao P. Vacant Parking Slot Detection in the Around View Image Based on Deep Learning. Sensors (Basel). 2020 Apr 10;20(7):2138.

[15] J. Redmon, S. Divvala, R. Girshick, and A. Farhadi, "You only look once: Unified, real-time object detection," in Proc. IEEE Int. Conf. Comput. Vis. Pattern Recognit., Jun. 2016, pp. 779–788.

[16] W. Liu, et al., "SSD: Single shot multibox detector," in Proc. Eur. Conf. Comput. Vis., 2016, pp. 21–37.

[17] J. Redmon and A. Farhadi, "Yolo9000: Better, faster, stronger," in Proc. IEEE Int. Conf. Comput. Vis. Pattern Recognit., Jul. 2017, pp. 7263–7271.

[18] Redmon, J.; Farhadi, A. Yolov3: An incremental improvement. arXiv 2018, arXiv:1804.02767.

[19] T. Qin, T. Chen, Y. Chen and Q. Su, "AVP-SLAM: Semantic Visual Mapping and Localization for Autonomous Vehicles in the Parking Lot," 2020 IEEE/RSJ International Conference on Intelligent Robots and Systems (IROS), Las Vegas, NV, USA, 2020, pp. 5939-5945.

[20] Scaramuzza, D., Martinelli, A. and Siegwart, R., (2006). "A Flexible Technique for Accurate Omnidirectional Camera Calibration and Structure from Motion", Proceedings of IEEE International Conference of Vision Systems (ICVS'06), New York, January 5-7, 2006.

[21] RidgeRun, RidgeRun's Birds Eye View project research, https://developer.ridgerun.com/wiki/index.php/Birds_Eye_View/Introduction/Research, Accessed 27 July 2023